\documentclass[10pt,twocolumn,letterpaper]{article}

\usepackage{cvpr}              

\usepackage[symbol]{footmisc}
\usepackage{graphicx}
\usepackage{amsmath}
\usepackage{amssymb}
\usepackage{booktabs}
\usepackage{subcaption}
\usepackage{multirow}
\usepackage{enumitem}
\usepackage{float}
\usepackage[accsupp]{axessibility} 

\usepackage[pagebackref,breaklinks,colorlinks]{hyperref}

\usepackage[capitalize]{cleveref}
\crefname{section}{Sec.}{Secs.}
\Crefname{section}{Section}{Sections}
\Crefname{table}{Table}{Tables}
\crefname{table}{Tab.}{Tabs.}

\def\cvprPaperID{10504} 
\def\confName{CVPR}
\def\confYear{2022}

\begin{document}

\title{TransRAC: Encoding Multi-scale Temporal Correlation with Transformers for \\
Repetitive Action  Counting}
\renewcommand{\thefootnote}{\fnsymbol{footnote}} 

\author{Huazhang Hu$^1$\footnote[1]{} ,   Sixun Dong$^1$\footnote[1]{} , Yiqun Zhao$^1$ , Dongze Lian$^{1,2}$ ,  Zhengxin Li$^1$\footnote[2]{} , Shenghua Gao$^{1,3,4}$\footnote[2]{}\\
$^1$ShanghaiTech University \qquad $^2$National University of Singapore\\
$^3$Shanghai Engineering Research Center of Intelligent Vision and Imaging\\
$^4$Shanghai Engineering Research Center of Energy Efficient and Custom AI IC\\
{\tt\small   \{huhzh, dongsx, v-zhaoyq, liandz,  lizhx, gaoshh\}@shanghaitech.edu.cn}\\
}
\maketitle

\footnotetext[1]{These authors contributed equally to this work.} 
\footnotetext[2]{Corresponding authors.} 

\renewcommand*{\thefootnote}{\arabic{footnote}}
\setcounter{footnote}{0}
\begin{abstract}
Counting repetitive actions are widely seen in human activities such as physical exercise. Existing methods focus on performing repetitive action counting in short videos, which is tough for dealing with longer videos in more realistic scenarios. In the data-driven era, the degradation of such generalization capability is mainly attributed to the lack of long video datasets. To complement this margin, we introduce a new large-scale repetitive action counting dataset covering a wide variety of video lengths, along with more realistic situations where action interruption or action inconsistencies occur in the video. Besides, we also provide a fine-grained annotation of the action cycles instead of just counting annotation along with a numerical value. 
Such a dataset contains 1,451 videos with about 20,000 annotations, which is more challenging. For repetitive action counting towards more realistic scenarios, we further propose encoding multi-scale temporal correlation with transformers that can take into account both performance and efficiency. Furthermore, with the help of fine-grained annotation of action cycles, we propose a density map regression-based method to predict the action period, which yields better performance with sufficient interpretability. Our proposed method outperforms state-of-the-art methods on all datasets and also achieves better performance on the unseen dataset without fine-tuning. The dataset and code are available \footnote{\url{https://svip-lab.github.io/dataset/RepCount_dataset.html}}.
\end{abstract}


\begin{figure*}[ht!]
  \centering
    \begin{subfigure}{0.48\textwidth}
      \centering   
      \includegraphics[width=1\linewidth]{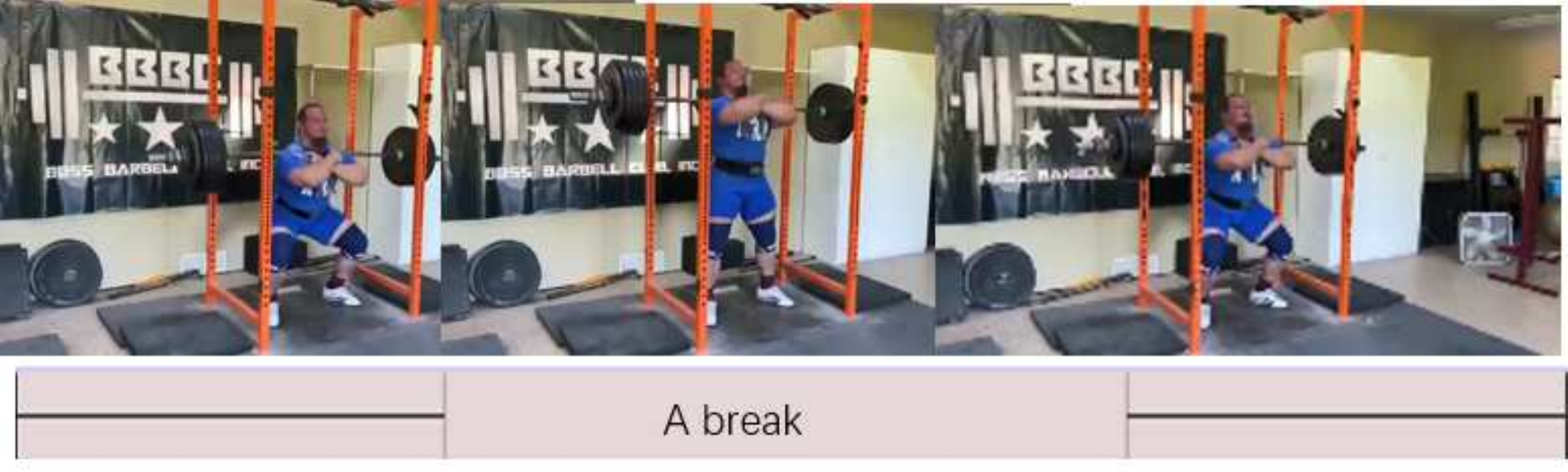}
        \caption{Interruption during the actions (squats)}
        \label{fig:1a}
    \end{subfigure}        
    \hfill  
    \begin{subfigure}{0.48\textwidth}
      \centering   
      \includegraphics[width=\linewidth]{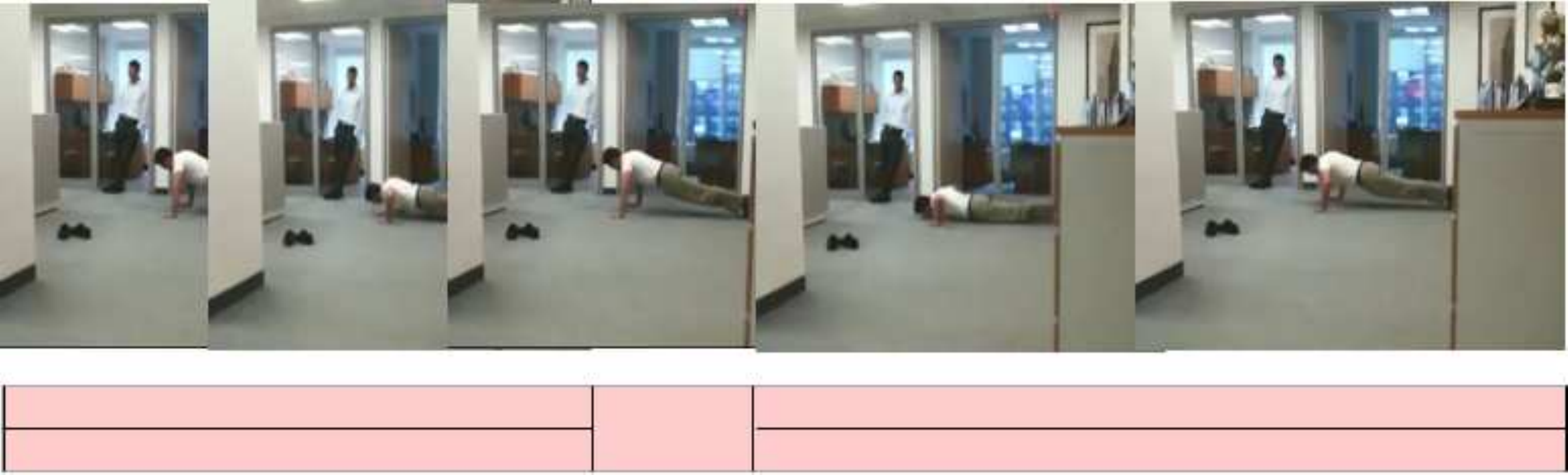}
        \caption{Inconsistent action cycles (push up)}
        \label{fig:1b}
    \end{subfigure}
    \hfill
    \begin{subfigure}{0.48\textwidth}
      \centering   
      \includegraphics[width=\linewidth]{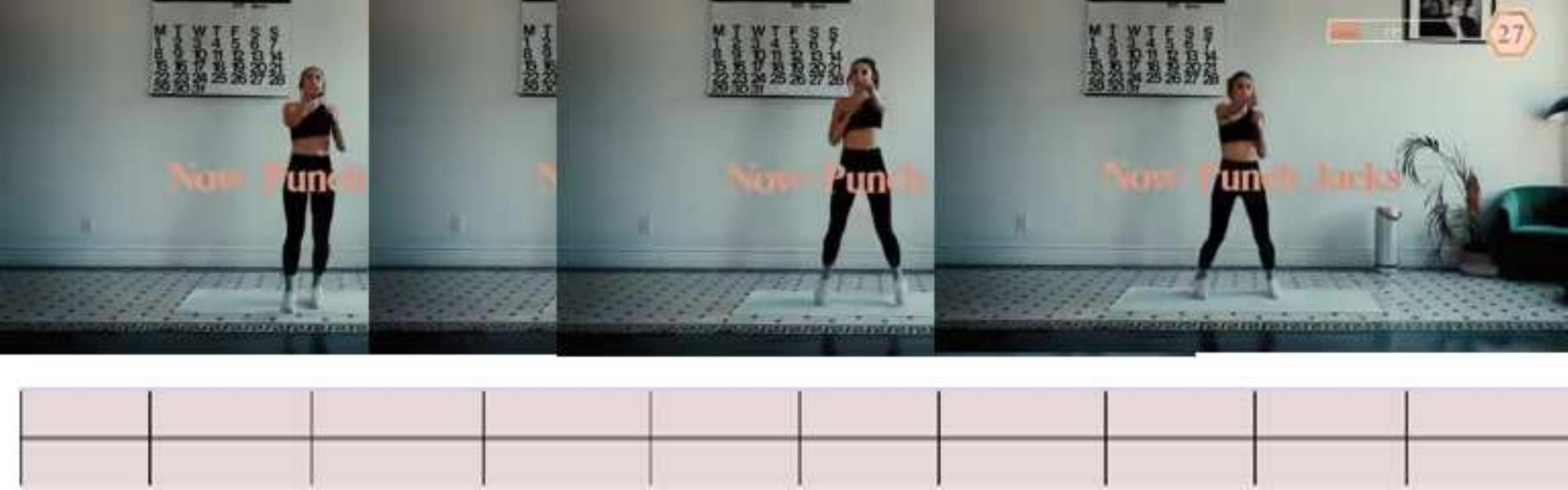}
        \caption{Long video with numerical cycles (60 seconds) (punch jacks)}
        \label{fig:1c}
    \end{subfigure}
    \hfill
    \begin{subfigure}{0.48\textwidth}
      \centering   
      \includegraphics[width=\linewidth]{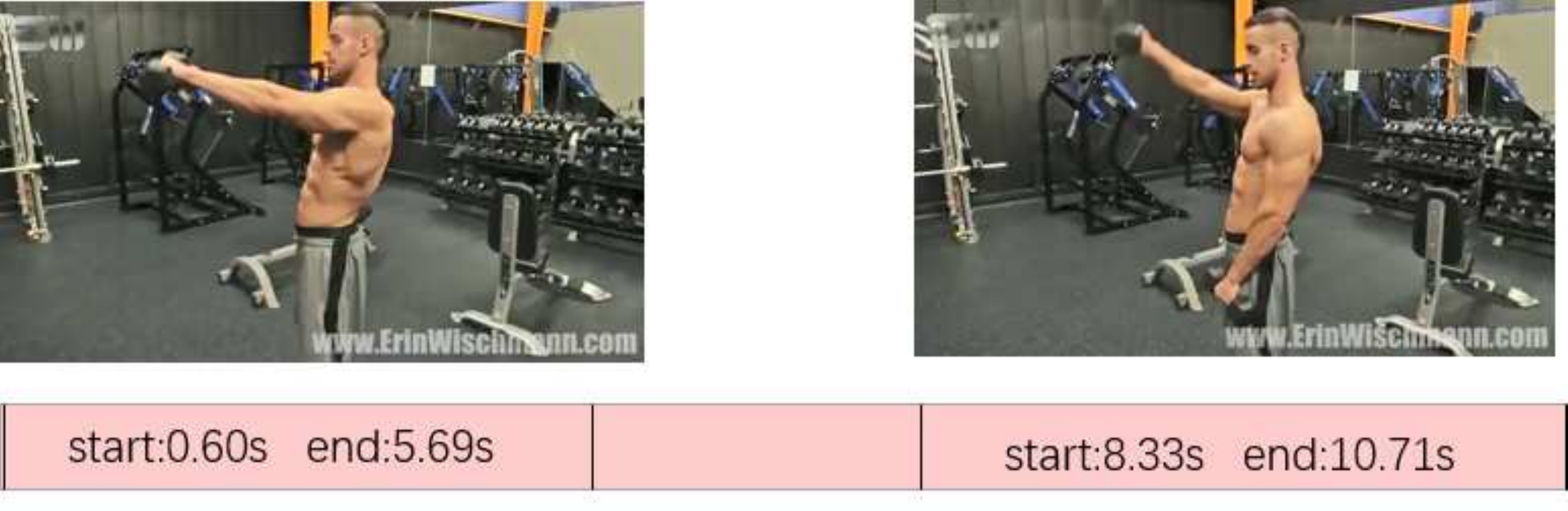}
        \caption{Annotations in the form of start and end of each cycle (front raise)}
        \label{fig:1d}
    \end{subfigure}
    
\caption{
The features of our proposed dataset \textbf{RepCount}: (a) anomaly case (interruption during actions); (b) anomaly case (inconsistent action cycles); (c) long video that consists of numerical action cycles; (d) the fine-grained labeling protocol. 
}
\label{fig:teaser}
\end{figure*}

\section{Introduction}\label{sec:intro}

%

Planetary motion, the change of seasons, and heartbeats, these periodic movements that are everywhere in our lives. They can be modeled by the Newtonian mechanics, or detected with the aid of sensors for the understanding of the world or our bodies. In computer vision, the detection of repetitive/periodic motions also plays an important role, such as in human activity, where the counting of some physical exercise movements benefits people in fitness detection and planning. Although one can use some sensors (\emph{e.g.}, gravity sensors) on the human body, vision-based approaches enable non-invasive and thus make third-view-based video analysis possible and promising. Repetitive action counting in computer vision is also useful as an auxiliary cue for other human-centric video analysis applications, such as pedestrian detection\cite{Pedestrian} and 3D reconstruction\cite{Li_2018_CVPR, Ribnick_3dreconstruction}.


Despite this importance, repetitive action counting methods in computer vision has rarely been explored. Previous papers \cite{Zhang_2020_CVPR,RepNet} tend to count repetitive actions in short videos, such as some simple videos grabbed from the Kinetics dataset \cite{kay2017kinetics}. However, these videos lack some realistic scenarios, which limits the application of the method in more realistic scenarios due to the following two points:

\begin{itemize}
\item \textbf{Restricted video length.} The previous datasets \cite{Runia_2018_CVPR,Zhang_2020_CVPR,RepNet,synthesized} typically contain only short videos (\emph{e.g.}, 0.4-30 s), however, methods are likely to be deployed to long videos in real scenarios. For instance, we count push-ups or jump-jacks with a video length of 60 s. Counting actions in such long videos is more challenging because there might exist various anomalies in real scenarios, such as the action being interrupted with internal or external reasons (\cref{fig:1a} ), or the inconsistency between action periods ( \cref{fig:1b} ). These anomalies might cause the previous algorithm to fail or obtain sub-optimal performance, affecting the generalization of the algorithm to real scenarios.

\item \textbf{Inadequate annotations.} In previous datasets \cite{Runia_2018_CVPR,Zhang_2020_CVPR,RepNet,synthesized}, the number of repetitive actions in a video is simply labeled as a numerical value. Although the count number serves as an ultimate predictive goal, such coarse-grained annotation deprives the interpretability of the algorithm. The model only predicts a numerical value during training or inference, which makes it difficult to evaluate the model more completely. As argued in some crowd counting papers \cite{MCNN,Wan_2019_ICCV,Ranjan_2018_ECCV,liu2019context}, the total number of repetitive actions is correct, but the position of the intermediate cycles might be wrong. 


\end{itemize}

\begin{figure*}[ht!]
    \centering
    \begin{subfigure}{0.5\textwidth}
      \centering   
      \includegraphics[width=1\linewidth]{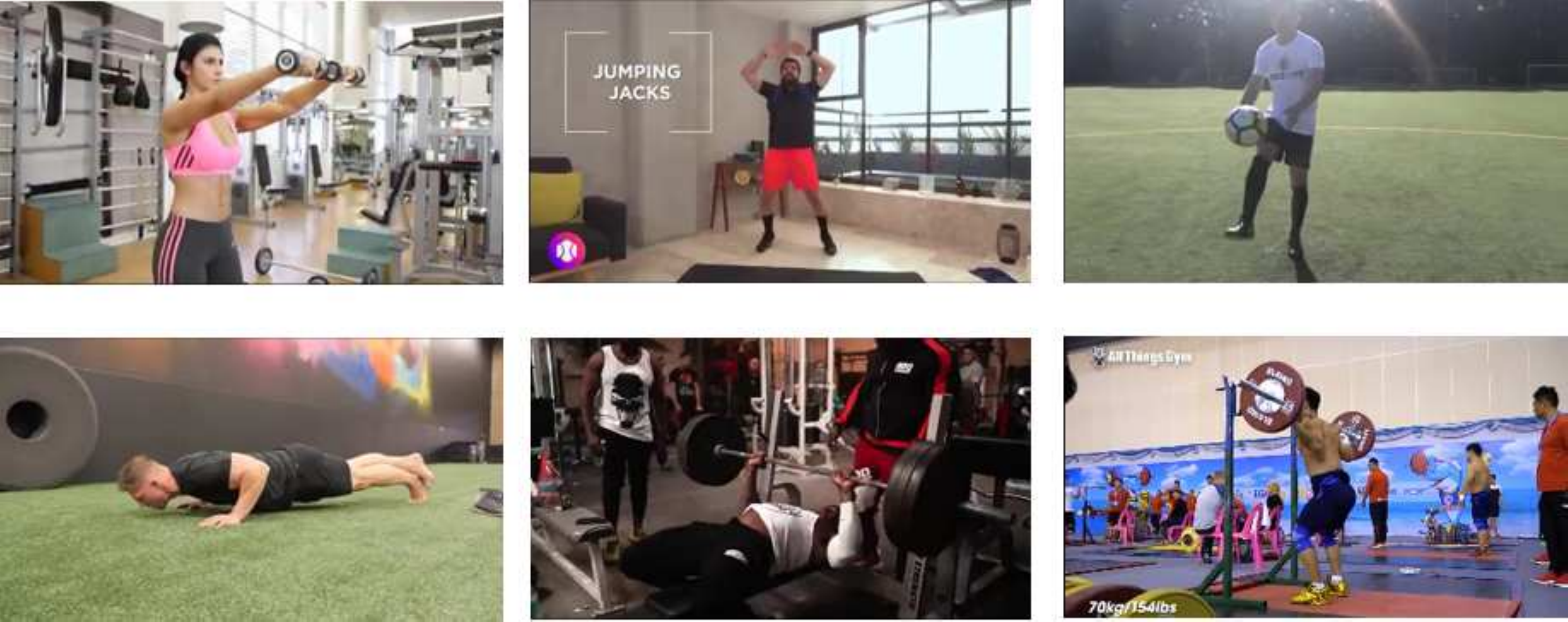}
       \caption{Dataset Part-A}
    \end{subfigure}      
    \hfill
    \vline
    \hfill  
    \begin{subfigure}{0.17\textwidth}
        \centering   
        \includegraphics[width=\linewidth]{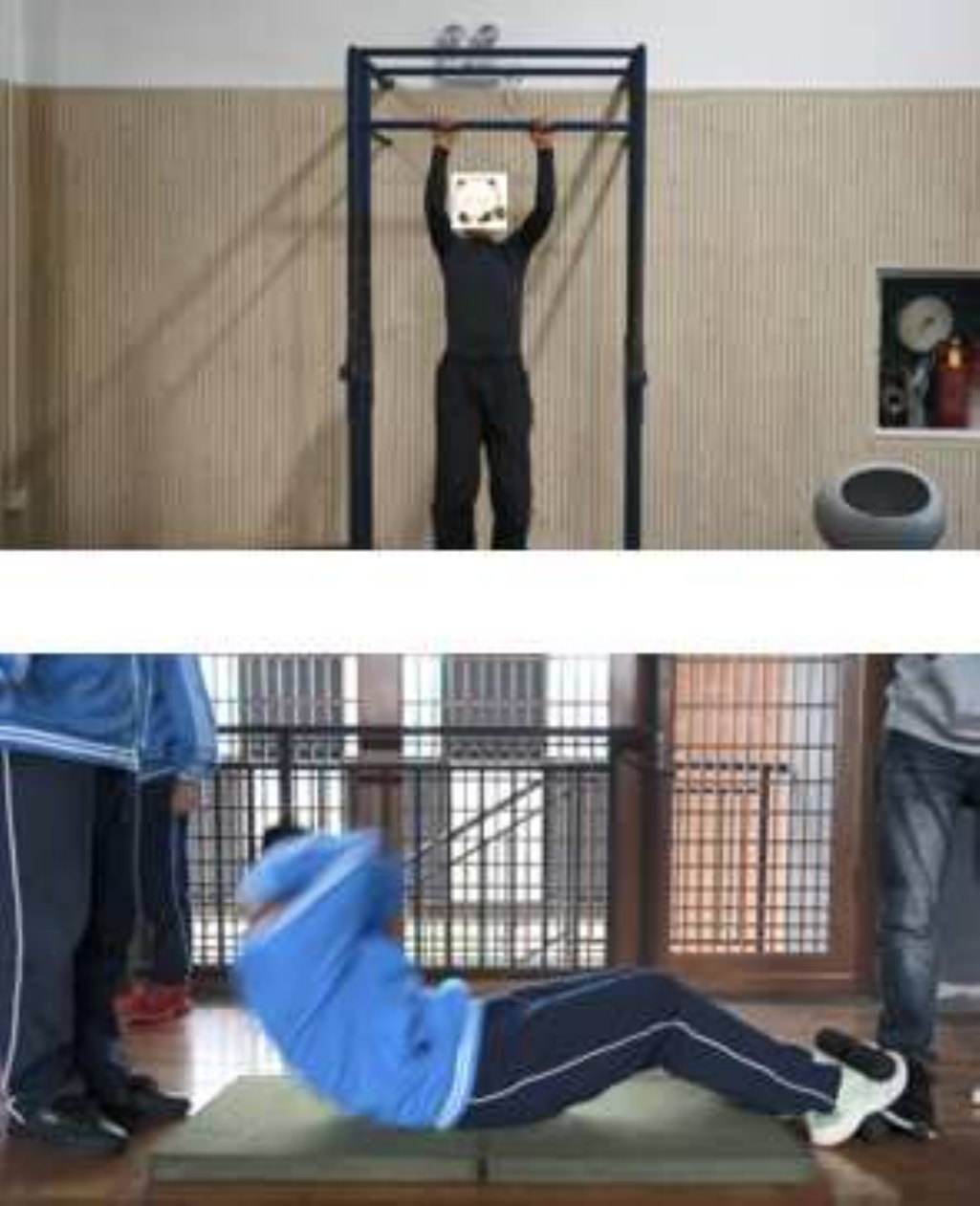}
        \caption{Dataset Part-B}
    \end{subfigure}
    \hfill
    \vline
    \hfill
    \begin{subfigure}{0.22\textwidth}
        \centering   
        \includegraphics[width=\linewidth]{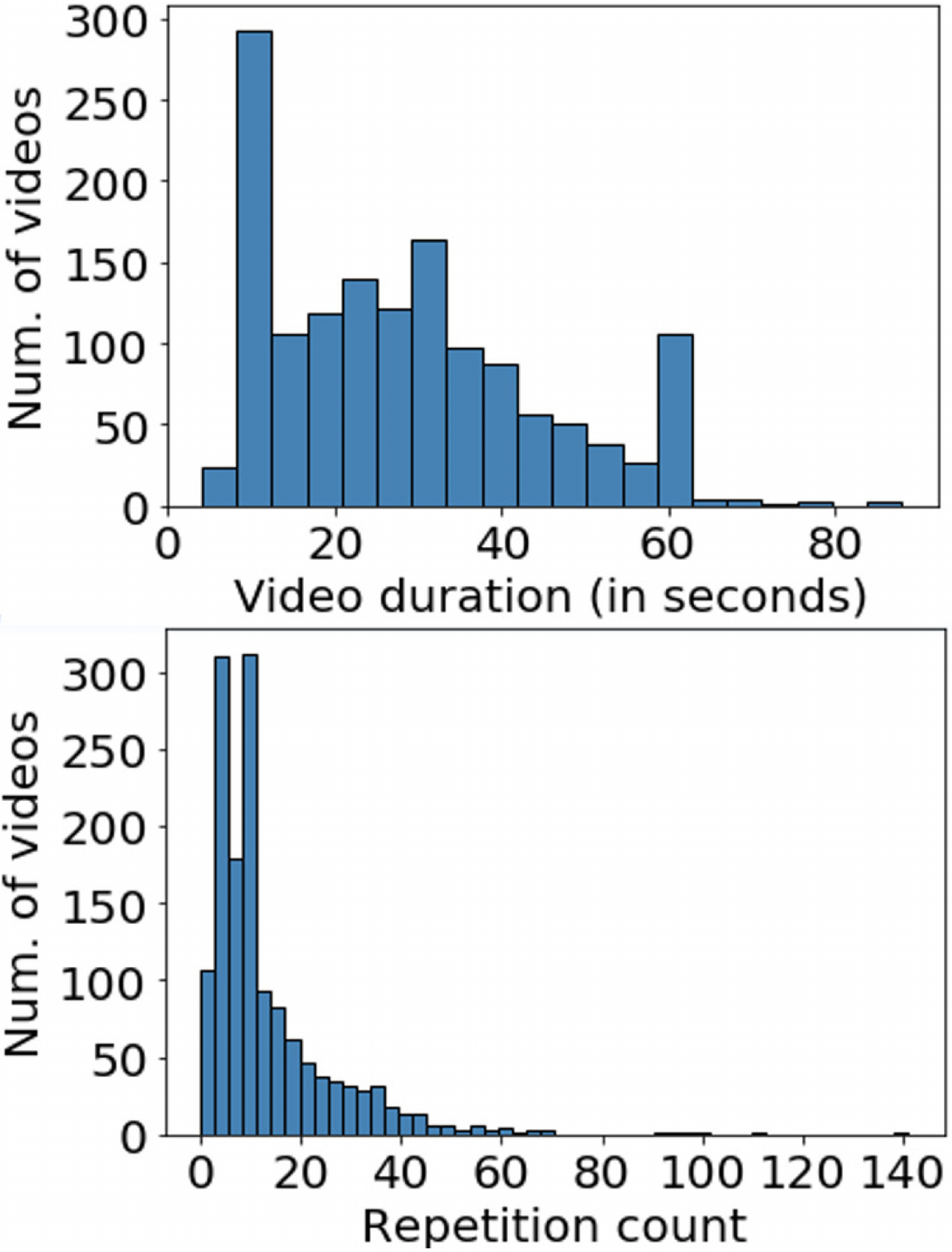}
        \caption{Data analysis}
    \end{subfigure}
    \caption{
    \textbf{The summary of proposed benchmark RepCount}: The first two columns represent the part-A and part-B respectively, the right column shows the statistics of video length and repetition count of our dataset.
    }
\label{fig:datashow}
\end{figure*}

In data-driven deep learning approaches, the dataset is the key to algorithmic innovation. To tackle the above problems, we collect a large-scale human-centric dataset, which is closer to the real one. As shown in \cref{fig:1c}, there are a large number of variations in video length, while the interruptions or inconsistent action cycles occur in some videos. For more accurate performance evaluation and model interpretability, we provide a more fine-grained annotation of the action cycles, such as \cref{fig:1d}. Further, we also collect a part containing student activity videos captured in a fully realistic scenario (in local school), which is significantly different from the previous datasets where the videos are crawled from YouTube. \cref{fig:datashow} provides an overview of our dataset. Such a dataset is more challenging and has the potential to become a new benchmark for repetitive action counting.

To perform repetitive action counting, previous methods \cite{RepNet} generally take a fixed number of frames for prediction. Such an approach might be reasonable in relatively short videos. For example, TSN \cite{TSN2016ECCV} extracts three frames for action recognition of trimmed videos, where the information characterizing the action is concentrated on certain keyframes. However, for long videos in the real scenario, extracting fixed frames will result in sub-optimal performance. Since the video duration varies very much (e.g. from 4s to 88s) and if the number of selected frames is too small, high-frequency actions will be neglected. On the contrary, if too many frames are selected, it might cause a waste of computational resources. Another alternative is to sample the video with the same frequency for both long and short videos. However, some actions are very fast (\eg, jumping rope) and some are very slow (\eg, push-ups). Sampling with a fixed frequency would either lead to performance degradation or would not be efficient enough. To balance performance and efficiency, we propose a multi-scale temporal correlation encoding network with transformers that can take care of not only high and low-frequency actions but also long and short videos. This approach allows the model to automatically select its adapted scale to compute the correlation matrix for final count prediction. Furthermore, thanks to the fine-grained annotation of action cycles in our dataset (see \cref{fig:1d} ), we also propose a density map regression-based method to predict action periods, which not only yields better performance but is also more beneficial for the interpretability of the model.

We summarize our contributions in three-fold:

\begin{itemize}[leftmargin=*]
\item We introduce a new dataset, named RepCount, which consists of 1,451 videos and about 20,000 fine-grained annotations. Such a dataset allows for a large number of video length variations and contains anomaly cases, thus is more challenging. 
\item A new multi-scale temporal correlation encoding network with transformers, which can take care of not only high and low frequency actions, but also long and short videos, is designed for repetitive action counting. 
\item The proposed method outperforms state-of-the-art methods on our proposed dataset and all other datasets. Furthermore, we also achieve better performance on the unseen dataset without fine-tuning.

\end{itemize}

\section{Related Works}
\label{sec:related}

\noindent \textbf{Temporal auto-correlation}. The temporal auto-correlation function is widely used in motion recognition\cite{kobayashi2009three,kobayashi2012motion,chen2017action} and human identification\cite{kobayashi2006three}. Auto-correlation in time series contains periodic information\cite{Auto-correlation1,Auto-correlation2}. The most common method to represent auto-correlation is the vector inner product. Vaswani et al.\cite{vaswani2017attention} obtain the auto-correlation matrices by multiplying \textit{query matrices} and \textit{key matrices}. Panagiotakis et al.\cite{Auto-correlation1} implement attention mechanism for video frame series to construct the auto-correlation matrices of video frames. With auto-correlation matrices, the cycle information from time series can be found to count the number of repetitive actions.

\noindent \textbf{Video feature extraction}.
For a long time, spatial and temporal convolution dominated the area of video feature extraction, such as C3D\cite{C3D}, I3D\cite{I3D}, P3D\cite{P3D}. However, limited by the small receptive field of the convolution kernel, a convolution-based method is hard to capture the long-range dependencies on the temporal domain. ViT\cite{ViT} and its variants bring a pioneering change for the Computer Vision field. However, Due to the quadratic  computational complexity and complex structure, training of the transformer-based model is costly \cite{kitaev2020reformer}. 
Video Swin Transformer\cite{video-swin-transformer} is a proper backbone due to its trade-off, which was pre-trained on a large dataset.

\noindent \textbf{Counting in computer vision}.
Counting from images or videos \cite{synthesized,Runia_2018_CVPR,Arteta16,lu2018class,lian2019density,Zhang_2020_CVPR,DBLP:journals/corr/abs-2103-13096,lian2021locating} is a very important field in Computer Vision. It has high application value in object detection\cite{seenouvong2016computer}, public transport\cite{lengvenis2013application} and physical exercises\cite{soro2019recognition}. Zhang et al.\cite{Zhang_2020_CVPR} propose a context-aware and scale-insensitive framework for temporal repetition counting. \cite{DBLP:journals/corr/abs-2103-13096} incorporate visual signal with corresponding sight signal to motion counting for the first time.

\noindent \textbf{Density map}.
The application of a density map enhances the effect of crowd counting \cite{MCNN,Wan_2019_ICCV,Ranjan_2018_ECCV,liu2019context}. The density map is generated from plot maps by convolving with a Gaussian kernel. The density map applies 2D  planar spatial distribution to represent the spatial distribution and the local probability distribution. \cite{MCNN} apply 2D density maps to achieve dense crowd counting. In \cite{tan2019crowd}, the extracted features are passed to the density regressor to generate a density map. In many neural network architectures, it could be regarded as an intermediate representation layer. A periodic density map preserves more information and gives the spatial distribution.

\noindent \textbf{Period annotation}.
Currently, data-driven learning methods have become an essential approach in computer vision. In the scenario of repetition counting, most datasets only label the period cycle count \cite{RepNet, Zhang_2020_CVPR,kay2017kinetics}. Researchers have to use generated data synthesized from real data and artificial data for training. \cite{synthesized} first proposed Synthetic data for the training model. However, such data are based on the assumption that the motion is continuous, uninterrupted, uniformly distributed, and with similar periods. \cite{RepNet} naively divided the count of periods by the number of frames to get the period length. This is far from the real motion situation. Therefore, a dataset with periodic fine-grained annotation is invaluable.


\begin{table*}
	\centering
	\resizebox{\linewidth}{!}{ 
		\begin{tabular}{@{}lccccc@{}}
			\toprule
			                         & \multicolumn{1}{c|}{{UCFRep}} & \multicolumn{1}{c|}{{Countix}} & \multicolumn{3}{c}{Ours}                                                                  \\
			\cline{4-6}
			                         & \multicolumn{1}{c|}{}          & \multicolumn{1}{c|}{}          & \multicolumn{1}{c|}{part-A} & \multicolumn{1}{c|}{part-B} & part-A + part-B               \\

			\hline
			Sources                  & Subset of UCF101               & Subset of Kinetics             & Youtube                     & Local school                & Collected by ourselves        \\
			Num. of videos           & 526                            & 8757                           & 1041                        & 410                         & 1451                          \\
			Duration Avg. $\pm$ Std. & 8.15 $\pm$ 4.29                & 6.13 $\pm$ 3.08                & 30.67 $\pm$ 17.54           & 28.53 $\pm$ 16.06           & \textbf{29.359$\pm$ 16.02}    \\
			Duration Min./Max        & 2.08/33.84                     & 0.2/10.0                       & 4.0/88.0                    & 5.56/79.16                  & \textbf{4.0/88.0}             \\
			Count Avg. $\pm$ Std.    & 6.66                           & 6.84$\pm$ 6.76                 & 14.99 $\pm$ 14.70           & 9.27 $\pm$ 4.36             & \textbf{ 15.932 $\pm$  15.65} \\
			Count Min./ Max          & 3/54                           & 2/73                           & 1/141                       & 1/32                        & \textbf{ 1/141}               \\

			\bottomrule
		\end{tabular}
	} 
	\caption{
		Dataset statistic of \textbf{Countix}\cite{RepNet}, \textbf{UCFRep}\cite{Zhang_2020_CVPR} and the proposed \textbf{RepCount}. Our dataset is lager than the previous datasets in terms of average duration and average annotations.
	} 
	\label{tab:dataset}
\end{table*}

\section{Our Proposed Dataset}
Existing repetition counting datasets, mainly including Countix \cite{RepNet} and UCFRep \cite{Zhang_2020_CVPR}, have been widely  consumed for the evaluation of repetition counting models. In these datasets, video clips that are collected from YouTube cover a variety of perspectives, dimension sizes and action categories. Typically, the total number of repetitive actions in a video clip is labeled as its ground truth. While these datasets significantly contributed to modeling the repetition counting problem, there still exist several non-negligible limitations that increase the gap between the scenarios illustrated in the videos and realistic ones, such as i) no interruption to actions, either from internal or external; ii) only containing uniform action frequency in an individual video; iii) the lack of long-range videos; iv) coarse-grained ground truth annotation, \emph{etc.} In particular, the last point hinders the development of more sophisticated models.

To overcome these limitations, we introduce a novel repetition counting dataset called \textbf{RepCount} that contains videos with significant variations in length and allows for multiple kinds of anomaly cases, as demonstrated in \cref{fig:datashow}. These video data collaborate with fine-grained annotations that indicate the beginning and end of each action period. Furthermore, the dataset consists of two subsets namely Part-A and Part-B. The videos in Part-A are fetched from YouTube, while the others in Part-B record simulated physical examinations by junior school students and teachers. Therefore, flexible strategies of data splitting for training and evaluation could be adopted according to the specific demand. Then we introduce the data collection, annotation and statistics in detail.  

\noindent \textbf{Dataset collection}. According to the source of data, our dataset consists of two parts. For part-A, we collected 1,041 video clips from YouTube. The type of actions includes workout activities (squatting, pulling-up, front-raising, \emph{etc.}), athletic events (rowing, pommel horse, \emph{etc.}) and other repetitive actions (soccer juggling). We select the video that represents at least one integral series of actions in line with human habits. Also, the videos usually contain some irrelevant actions like speaking and relaxation. For the most important, the interruption during an action series is preferable, which may result in difficulties for accurate counting. We use the open-source script \textit{YouTube-download}\footnote{\url{ https://ytdl-org.github.io/youtube-dl/index.html}} to download the videos and edit them to keep only useful clips. The length of each video clip is 20-40 seconds in major. For part B, we record the videos of exercises such as sitting- and pulling-up done by volunteers. 

\noindent \textbf{Dataset annotation}. The existing repetition counting datasets simplify the problem by assuming that the actions are periodically uniform and not interrupted by irrelevant situations. Thus they have only coarse-grained annotations in the form of a single-valued total \cite{RepNet} or two timestamps that indicating the start and end time of the whole action \cite{Zhang_2020_CVPR}. Our protocol for fine-grained annotation is as follows: i) each individual video is assigned to two volunteers; ii) start and end time of every action cycle are labeled; iii) the annotations are cross-validated by comparing that from two volunteers, which should be inspected and revised if they differ more than 1 in total. Following the protocol, each movement cycle is precisely positioned on the time axis, which enables the design and training for models with better interpretability.

\noindent \textbf{Dataset statistics}. The summary of our dataset is shown in \cref{tab:dataset}. In brief, we provide 1,451 videos collaborated with 19,280 annotations. The videos from our dataset have an average length of 39.359 seconds, which is 4-5 times the length of videos from other datasets. Each video clip in our dataset contains 15.932 action cycles in average, while that is 6.66 for \emph{UCFRep} and 6.84 for \emph{Countix}. Furthermore, part-B is constructed for the validation of model generalization. The graphical statistics is shown in the right part of \cref{fig:datashow}. Our dataset is featured with more realistic scenarios and fine-grained annotations.

\section{TransRAC Model}

\begin{figure*}
\centering
\centerline{\includegraphics[width=\textwidth]{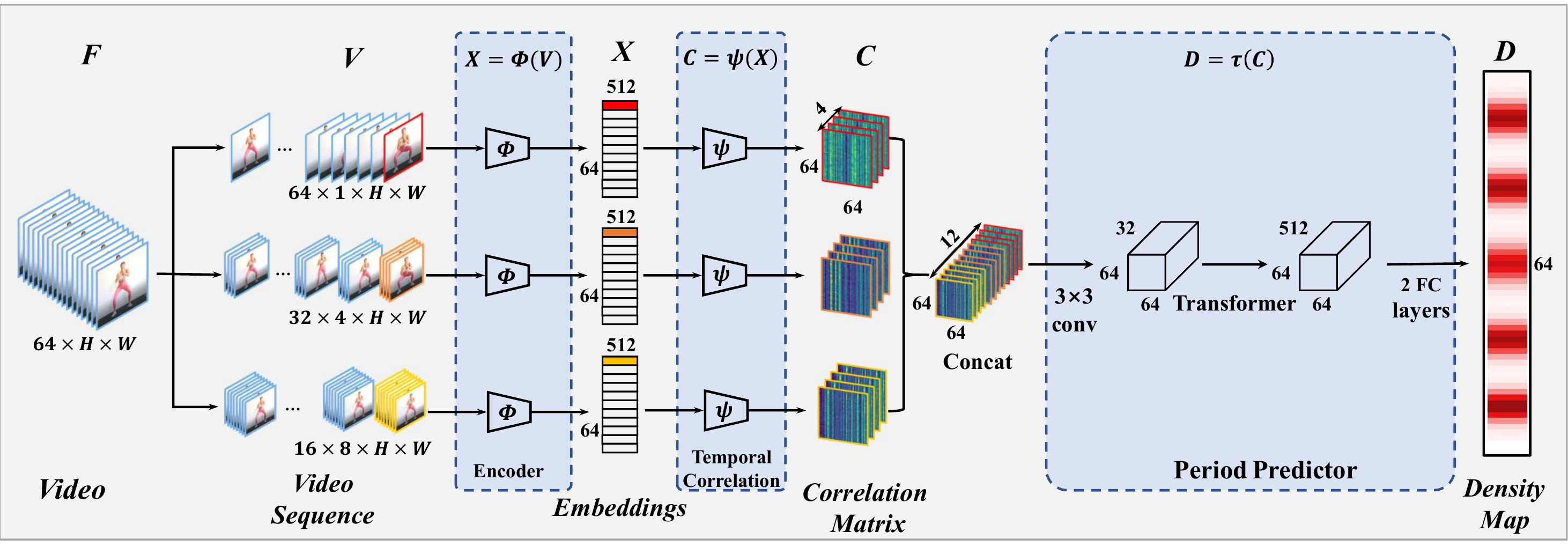}}
\caption{
\textbf{TransRAC architecture.}
We used three sliding windows with step sizes 1, 2, and 4 to generate the video sequence with an overlap: red, orange, and yellow.
Then extract features from multi-scale video sequences by the encoder. Calculate the correlation matrix in three scales, respectively. After concatenating three correlation matrices into one, make it throughout the remaining network and output the final density map.}
\label{fig:overview}
\end{figure*}

\label{sec:TransRAC Model}
Given a long-duration video that has more than 15 repetitive activities happening in the content, our goal is to count the number of repetitive actions. To achieve this, we propose the model called TransRAC that contains three stages: the encoder, temporal correlation, and period predictor. The video subsequences $V$ are fed into the encoder, then the output $X$ is used to calculate the correlation matrix $C$ by $C=\phi(X)$. At last, using period predictor $D=\tau(C)$ predict the final output density map.
\subsection{Encoder}

As shown in \cref{fig:overview}, the encoder is $X = \phi(V)$. In order to explain the function $\phi$, firstly, assuming that we have a sequence of $N$ frames $F = [f_1,f_2,\dots,f_N]$. we extract the three scale video  subsequence $V$ from the original as the input of $\phi$. Then we feed the video sequence $V$ into the encoder $\phi$ to produce multi-scale embeddings $X = [X_1,X_4,X_8]^T$, where $X_1 = [x_1^1,x_1^2,\dots,x_1^N]^T$. $X_4$ and $X_8 $ are similar to $X_1$. 

\noindent \textbf{Video sequences of multi-scale}. We extract three scale subsequences from the video: single-frame, 4-frames, and 8 frames indicating the video subsequence of multi-scale (V) in \cref{fig:overview}. As \cref{eq:V_148}, $V_1, V_4, V_8$ indicates the element length of the V. In the video sampling, we use a sliding window with a step size of 2 to obtain $V_4$ and a step size of 4 to get $V_8$. And we also need to pad the video to ensure the same temporal dimension of output three scale sequences.
\begin{equation}
\left \{ 
\begin{array}{c}
    V_1 = \{\{f_1\};\{f_2\};\dots\}    \\
    V_4 =\{\{f_1,\dots,f_4\};\{f_3,\dots,f_6\};\dots\} \\
    V_8 = \{\{f_1,\dots,f_8\};\{f_5,\dots,f_{12}\};\dots\} 
\end{array}
\right.
\label{eq:V_148}
\end{equation}


\noindent \textbf{Spatio-temporal features}. The video swin transformer\cite{video-swin-transformer} is  used to extract 3d features from individual video subsequences of different scales. It can easily capture the long-range dependencies using the self-attention mechanism; at the same time, with the hierarchical design, it can also capture the local dependencies, which is more suitable for the image.

Let video subsequence $V_i$, where $i\in \{1,4,8\}$, pass through the feature extraction block to extract the features. Three kinds of video clips with different scales can  match different period lengths (\emph{e.g.} jump jacks and squat) better. The resulting features of each scale are of size $ 7\times{7} \times {t} \times{768}$, where the $t$ is equal to a 2-fold compression in the temporal dimension. Then all these feature were concatenated in temporal dimension as a feature block . 

\noindent \textbf{Temporal context}.
To take into account more temporal context, we apply a layer of 3D convolution after the feature extractor, which has $3\times3\times512$ filters with ReLU activation. After that, we use a Global 3D Max-pooling layer over the spatial dimensions to reduce model parameters and obtain the final result $X$ of encoder $\phi$ as the embeddings in \cref{fig:overview}. The above operations are carried on different scale video subsequences to the extent that we can obtain more information on the time domain.

\subsection{Temporal correlation and Self-attention}
The correlation of embeddings can be expressed as $C_i = \Psi(X_i)$. We need to compute every correlation $c_p^
{i}$ between $x_p^i$ embeddings with other $x_p^j$, where $j \in \{1,2,\dots,N\}$ and $j \neq i$, such that we can use the embedding $X_p$ to obtain the correlation matrix $C_p = [c_p^1,c_p^2,\dots,c_p^H]^T$, $p\in\{1,4,8\}$ and $H$ is the number of attention-head.

\noindent \textbf{Correlation matrix}.
For the temporal locations of the activities, we use transformers\cite{vaswani2017attention} with correlation-matrix and self-attention mechanism to encode multi-scale temporal correlation layers. After encoder the video sequences, we can get embeddings $X_i$ for each scale $V_i$, where $i\in{1,4,8}$. And the shape of all embeddings for every scale is $64\times512$ as \cref{fig:overview}. Then We use the self-attention mechanism to calculate the correlation matrix. One scale embeddings $X_i$ is multiplied with two matrices of weights for obtaining keys matrix called $K$ and query matrix called $Q$. Then we could use $K$ and $Q$ to calculate attention scores, which should be called correlation in this paper.

\noindent \textbf{Self-attention}.
We construct the correlation matrix C by $C = f(Q,K)$, where $f(.)$ is known as dot product attention. And as \cref{fig:overview} showed, these are two more important points that We use 4-heads with 512 dimensions (not eight heads which is more usual \cite{vaswani2017attention}) and multi-scale embeddings to calculate the correlation. Therefore, after the self-attention layer, concatenating three scales' features into one, we could get the shape of output is $[N, N, M\times H]$. M means how many scales we have. $ H$ and $N$ are the numbers of heads and input frames independently. Furthermore in details, $N$, $M$ and $H$ in TransRAC are 64, 3 and 4, respectively.

\subsection{Period Predictor}
In \cref{fig:overview}, $D=\tau(C) $ shows feed $C$, where $C$ is concatenated from $C_1$, $C_2$ and $C_3$, to the density map predictor which outputs none element for each video subsequence : the value of density  the $D = [d_1,d_2,\dots,d_N]$ represents the distribution of period. A more detailed version can be seen in \cref{fig:overview}.

\noindent \textbf{Density map}.
The most straightforward advantage of the approach of density map is that it has a strong ability for explanation.. Therefore, We use the density map predicto as our period predictor. The density map contains the global information of the entire video. Each row of the density map indicates the frame's position in the local cycle and the distribution of the frame in the global video. We also compared density map regressor with classifiers in ablation experiments and found density maps performed. And more details for comparisons between the two predictors can be seen in \cref{sec:experiment}. For more implementation details can be seen in the supplementary material.

 \subsection{Losses}  
 Our dataset, \emph{RepCount}, is annotated with each position of each motion period in the temporal dimension. Pass those labels through the Gaussian function $G(x)$ to get the ground truth. The process of Gaussianization can be seen in the supplementary material. Therefore, using MSE (Mean Squared Error) as the loss function tends to be a good choice.

\subsection{Inference}
To compare the network's performance purely from an academic view, we have not taken any measures to improve the prediction accuracy which is used in previous work\cite{RepNet}. The way to infer the counting of  repetitions has the following operations:

\noindent \textbf{Sample video}.
For a video with any length of fewer than two minutes, we directly sample 64 frames. If the input video has less than 64 frames, we will implement padding in the temporal domain.
       
\noindent \textbf{Calculating}.
These frames are input into the model to obtain the prediction results of density map $D=[d_1,d_2,\dots,d_N]$. Applying a linear sum to obtain the predicted value $\hat{p}$ of the number of action periods, where $d_i$ means the value of density map.

\section{Experiments}
\label{sec:experiment}

 There are five central parts in this section. First of all, we explain some existing benchmarks and the evaluation matrices used in popular repetition counting. Secondly, We illustrate the advantages and capabilities of fine-grained annotations in detail. By visualizing and comparing the predictions for different sports, we propose conjectures and solutions. Then we evaluate our model performance and compare it to other methods, which were trained on our dataset \emph{RepCount}, on the existing benchmarks. At last, we make an ablation study to justify our model design.

\subsection{Benchmarks and Evaluation Matrices}
We evaluate our method on the four video datasets: our test set of \emph{RepCount part-A}, ours \emph{RepCount part-B} and \emph{UCF Rep}\cite{Zhang_2020_CVPR}. As illustrated in \cref{tab:dataset}, \emph{ Ours (part-A+part-B)} contains videos with more count and longer duration than all  existing datasets. The previous work\cite{RepNet, Zhang_2020_CVPR} mainly uses two  matrices for evaluating repetition counting in videos:

\noindent \textbf{Off-By-One (OBO) count error}. If the predicted count is within one count of the ground truth, we can consider this video are counted correctly. Otherwise, it is a situation of counting error. It represents the error rate of repetition count over the entire dataset.

\noindent \textbf{Mean Absolute Error}. This metric means normalized absolute error between the ground truth count and the predicted count.
OBO and MAE are defined as follows:
\begin{equation}
\text{OBO}=\frac{1}{N}\sum_{i=1}^{N}[| \widetilde{c_i} - {c_i}|\leq{1}],
\end{equation}

\begin{equation}
\text{MAE}=\frac{1}{N} \sum_{i=1}^{N}\frac{|\widetilde{c_i}-{c_i}|}{\widetilde{c_i}},
\end{equation} 
where $\widetilde{c} $ is the ground truth repetition counts. \emph{N} is the number of given videos.

\subsection{Implementation Details}
We implement our method with PyTorch. The encoder, Video Swin Transformer tiny\cite{video-swin-transformer}, was pre-trained on the \emph{Kinetics400}. Using three columns to obtain input video sequences and feeding them into the encoder. Then we apply 2D convolution to fuse multi-scaled correlation matrix. The hidden layer dimension of transformer-based period predictor is 512. Limited by the memory of GPU, the parameters of the pre-trained encoder were frozen during the training process. We train our remaining layers of the model for 16K steps with a declining learning rate of $8\times 10^{-6}$ and optimized by the Adam optimizer using a batch size of 16. Additional details are provided on the code.

\subsection{Fine-grained Annotation}
\begin{figure}[ht]
\centering

\centerline{\includegraphics[width=1\columnwidth]{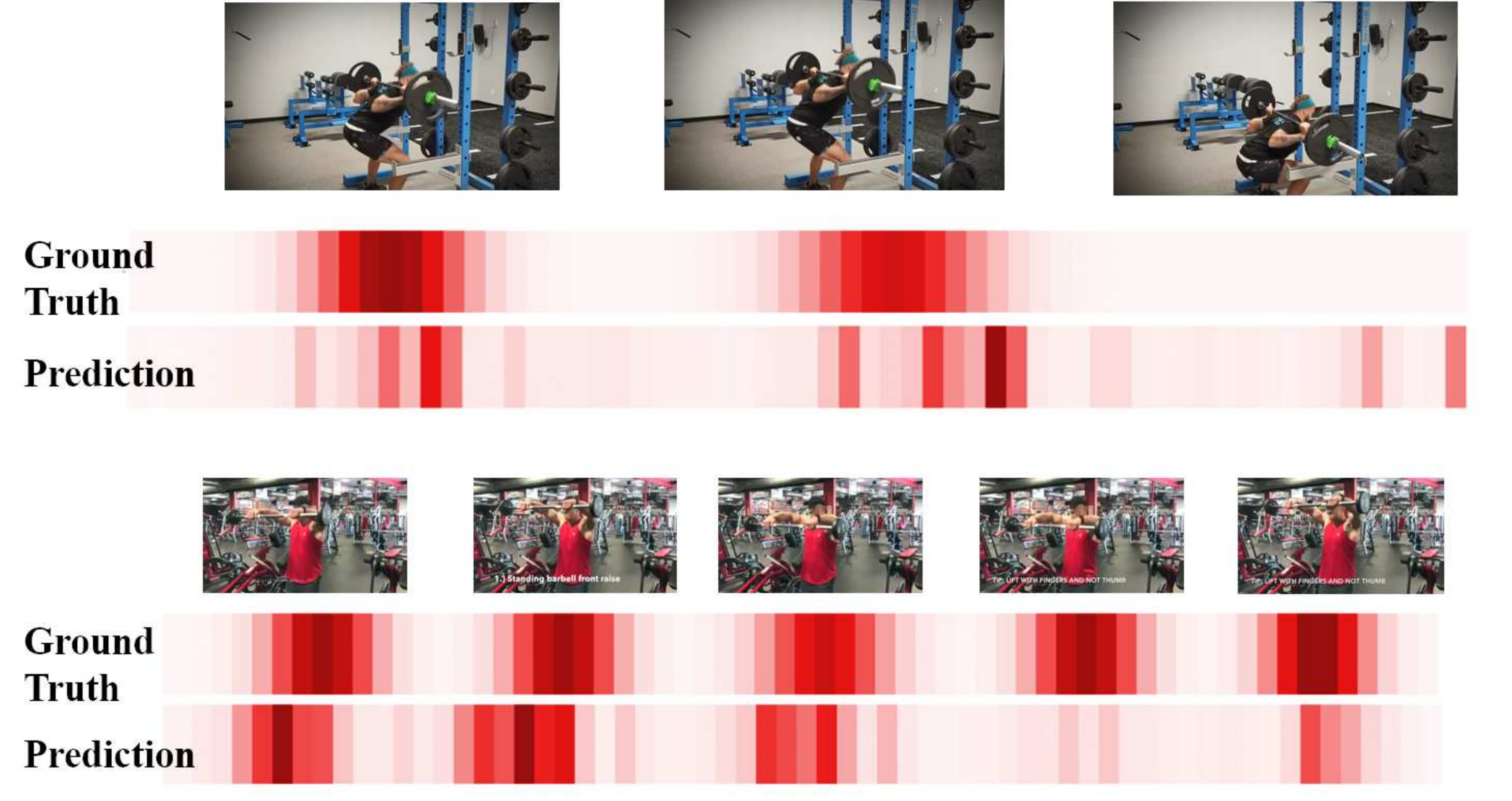}}
\caption{
\textbf{Visualization of density map.}
Here are comparisons between the ground truth and prediction result from our model. We can see from the first pair that the duration of videos in our dataset varies.  }
\label{fig:visual}
\end{figure}

Observing from the \cref{fig:visual}, it is easy to find accurate periodic location information on the ground truth, which is essential for accurately counting. As each kind of action has different characteristics, some activities, such as bench press, will be completed speed rate highly, due to the excellent energy of people at the beginning. However, at the end of the action, the rate will slow down. On the other hand, the period length of specific actions can be more uniform. As the examples of \cref{fig:visual} shown, "front-raise" is easier for the man to finish in a statable period. It is because we annotated the data in a more fine-grained way, by which we can get the locations of all kinds of actions from our dataset, we have the opportunity to fine-tune the structure of the model for different needs. Of course, there is no chance to set the density map as a predictor of our model without fine-grained annotations. Overall, a more fine-grained annotation is necessary to precisely help the model count the number of periods.  

\subsection{Evaluation and Comparison}
We evaluate the effectiveness of the model from multiple aspects. When we compare the TransRAC proposed with RepNet on \emph{RepCount (Part-A and Part-B)} and \emph{UCFRep} datasets, for a fair comparison, we modify the last fully-connection layer of RepNet\cite{RepNet} to make it capable of handling those videos containing more than 32 action periods. Unless otherwise specified, we train the networks on \emph{RepCount Part-A} and validate them on the test set of \emph{Part-A}, obtaining the results shown in \cref{tab:RepCount}. In addition, we compare some SOTA action recognition methods \cite{X3D,TANet,video-swin-transformer} and change output layers accordingly to adapt to our task. 
Farther more, we also compare the SOTA method \cite{huang2020improving} in the action segmentation field. More detail can be seen in the supplementary materials. One can observe that TransRAC, our model outperforms them by a notable margin on all the considered datasets.

\noindent \textbf{Generalization}.
From \cref{tab:B+UCF}, it also can be seen that the TransRAC model generalizes well on multiple datasets. 

\begin{table}[ht]
	\centering
	\resizebox{0.65\linewidth}{!}{
		\begin{tabular}{c|cccc}
			\toprule
			\centering
			                                          & \multicolumn{2}{c}{RepCount A}                          \\ \hline
			Method                                    & \multicolumn{1}{c|}{MAE$\downarrow$} & OBO $\uparrow$   \\ \hline
			X3D \cite{X3D}                            & \multicolumn{1}{c|}{0.9105}          & 0.1059           \\
			TANet \cite{TANet}                        & \multicolumn{1}{c|}{0.6624}          & 0.0993           \\
			Video SwinT \cite{video-swin-transformer} & \multicolumn{1}{c|}{0.5756}          & 0.1324           \\ 
			Huang et al. \cite{huang2020improving}    & \multicolumn{1}{c|}{0.5267}          & 0.1589           \\ \hline
			RepNet \cite{RepNet}                      & \multicolumn{1}{c|}{0.9950}          & 0.0134           \\
			Zhang et al.\cite{Zhang_2020_CVPR}        & \multicolumn{1}{c|}{0.8786}          & 0.1554           \\
			\textbf{Ours}                             & \multicolumn{1}{c|}{\textbf{0.4431}} & \textbf{0.2913 } \\ \hline
		\end{tabular}
	}

	\caption{
		Performance of different method on \emph{RepCount part-A test} when trained on the same train set of \emph{RepCount}.
	}
	\label{tab:RepCount}
\end{table}

\begin{table}[ht]
	\centering
	\begin{tabular}{c|c|c|c|c}
		\hline

		                     & \multicolumn{2}{c|}{RepCount B} & \multicolumn{2}{c}{UCFRep}                                         \\
		\hline
		Method               & MAE  $\downarrow$               & OBO  $\uparrow$            & MAE $\downarrow$     & OBO $\uparrow$ \\
		\hline
		RepNet \cite{RepNet} & 0.9994                          & 0.0025                     & 0.9985               & 0.009          \\
		Ours                 & 0.7839                          & 0.091                     & 0.6401               & 0.324          \\
		\hline
		\multicolumn{1}{c}{} & \multicolumn{1}{c}{}            & \multicolumn{1}{c}{}       & \multicolumn{1}{c}{} &
	\end{tabular}
	\caption{
		Performance of different method on \emph{RepCount part-B} and \emph{UCFRep} when trained on the same train set of \emph{RepCount part-A}.
	}
	\label{tab:B+UCF}

\end{table}

\begin{figure}[ht]
\centering

\centerline{\includegraphics[width=1\columnwidth]{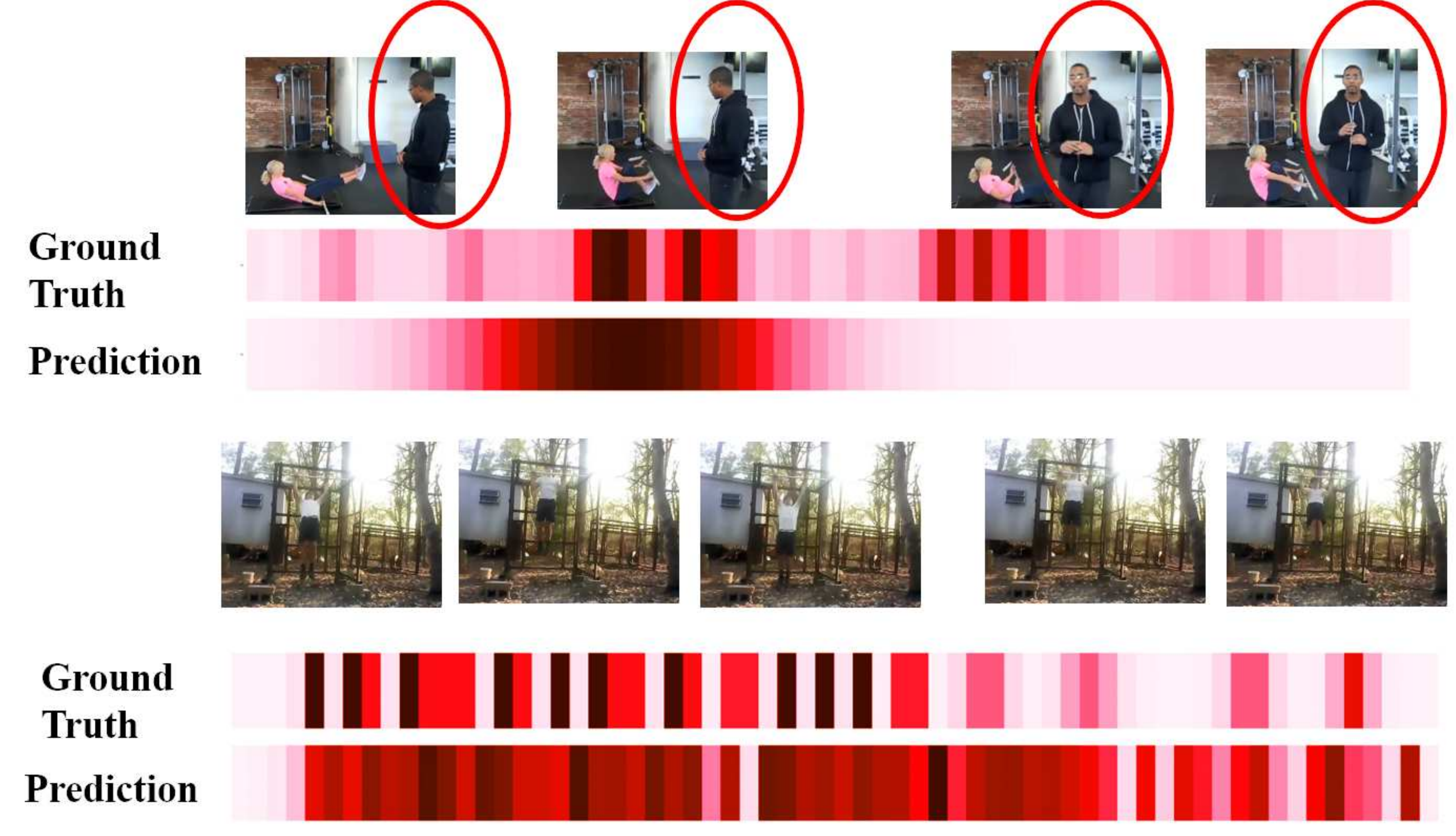}}
\caption{
\textbf{Visualization of bad cases.} Here are  results of two bad cases our model predicted. In the first case, another people is moving}
\label{fig:badcase}
\end{figure}
While our TransRAC performs well on the major part of the data, there still are some failure cases, as \cref{fig:badcase} shown. For the top one in \cref{fig:badcase}, we achieve a bad predicted result because there is more than one person  moving in the video. The failure case on the bottom indicates that the frame extracting strategy could diminish the performance of our model in some extreme situations. It can be seen that there is an apparent difference between the predicted density map and the ground truth, especially in the left part. As this sample video has a total number of frames of 772, most of the actions concentrate in the first 400 frames, neither the ground truth density map nor the output of our model is capable of handling such imbalance.

\subsection{Ablation Studies}

We perform several ablations to justify the decisions made while designing TransRAC. We train our model on a train set of \emph{part-A} and then evaluate the model on the test set of \emph{part-A}. More ablation experiments can be seen in the supplementary material.

\noindent \textbf{Correlation matrix}. In \cref{tab:TSM}, We compare the impact of applying different correlation matrix to our model. Temporal self-similarity matrix (TSM) \cite{RepNet} applies squared euclidean distance as the similarity function. But we found using the self-attention mechanism to calculate the correlation matrix is better. Since the experiment illustrates the self-attention mechanism could substantially improve the performance of our model, our model uses the self-attention mechanism.

\begin{table}[ht]
\centering
\begin{tabular}{c|c|cc} 
\cline{1-3}
               & \multicolumn{2}{c}{RepCount A} &   \\ 
\cline{1-3}
Correlation matrix  & MAE $\downarrow$  & OBO $\uparrow$ &   \\ 
\cline{1-3}
TSM          & 0.5678 & 0.2251                &   \\
Self-attention (Ours) & 0.4431 & 0.2913                &   \\
\cline{1-3}
\end{tabular}
\caption{
Result of our model applying different correlation matrix when trained on training set of \emph{RepCount part-A}.
}
\label{tab:TSM}
\end{table}

\noindent \textbf{Density map}. We build four models to verify the effectiveness of the density map, as shown in the first four rows of \cref{tab:ablation}. We could conclude that using density map regressor as the period predictor is significantly better than original classifiers. As shown in the third and the four rows of \cref{tab:ablation}, if we replace the classifiers with density map regressor, RepNet's performance has been significantly improved. The comparison result indicates that the Density map is more suitable for repetitive action counting. 

\begin{table}[ht]
	\begin{center}
		\begin{tabular}{c|cc}
			\hline
			                                    & \multicolumn{2}{c}{RepCount A}                         \\ \cline{1-3}
			Method                              & \multicolumn{1}{c|}{MAE$\downarrow$} & OBO  $\uparrow$ \\ \cline{1-3}
			ResNet  \cite{ResNet} + CLS          & \multicolumn{1}{c|}{0.9950}          & 0.0134          \\
			ResNet  \cite{ResNet} + DM           & \multicolumn{1}{c|}{0.6905}          & 0.0811          \\
			SwinT \cite{swin-transformer} + CLS & \multicolumn{1}{c|}{0.7027}          & 0.118           \\
			SwinT \cite{swin-transformer} + DM  & \multicolumn{1}{c|}{0.6781}          & 0.138           \\
			Ours (Scale-1)                             & \multicolumn{1}{c|}{0.6595} & 0.1854                  \\
            Ours (Scale-4)                             & \multicolumn{1}{c|}{0.5434} & 0.2649                                  \\
            Ours (Scale-8)   & \multicolumn{1}{c|}{0.6657} & 0.192  \\
			Ours (Multi)                       & \multicolumn{1}{c|}{\textbf{0.4431}} & \textbf{0.2913} \\ \cline{1-3}
		\end{tabular}
	\end{center}
	\caption{
		The result of abliation study when models trained on \emph{RepCount part-A}.
		ResNet+CLS indicates the original structure of \textbf{RepNet}\cite{RepNet}. ResNet + DM indicates replacing the last layers with the density map regressor. The same applies to swinT which indicates swin-transformer. Ours (Scale-X) indicates the single column without multi-scale correlation, where X represent $V_1$, $V_2$, and $V_3$. Ours (Multi) indicates our proposed structure.
	}

	\label{tab:ablation}

\end{table}

\noindent \textbf{Multi-scale}. In \cref{tab:ablation}, we compare the impact of applying different scales. We find that the multi-scale model performs better than the single-scale model when the number of frames is equal. The experiment demonstrates that more temporal features at different scales can obtain more period information. It is evident that multi-scale fused method brings a considerable benefit for model. 

\section{Conclusion}

In this paper, considering the tough problems of existing methods in dealing with long videos in more realistic scenarios, we propose a new large-scale repetitive action counting dataset. Such a dataset covers a wide variety of video lengths where action interruption or action inconsistencies situations occur in the video, this is more realistic. For model interpretability and more accurate evaluation, we further provide fine-grained annotation. The overall dataset contains 1,451 videos with about 20,000 annotations, which is more challenging and has the potential to be a new benchmark.
To balance the performance and efficiency, we propose to encode multi-scale temporal correlation with a transformer to tackle the repetitive action counting problems in realistic scenarios. We also propose a density map regression-based method to predict the action period, which yields better performance with sufficient interpretability. Extensive experiments show that our method achieves state-of-the-art results on all datasets and also achieves better performance on the unseen dataset without fine-tuning.

\noindent \textbf{Broader Impact and Limitations.} The proposed dataset is about the counting of repetitive actions, which means videos from our dataset are human-centric. The abuse of our dataset may cause privacy leaks. The usage of our dataset is limited to academic research. The proposed method predicts results based on the dataset, which may include some negative social impacts. Thus, the results conducted by our method may reflect the bias from the dataset. Other technical limitations are talked about in the \cref{sec:experiment}.

\noindent \textbf{Acknowledgements}. 
The work was supported by National Key R\&D Program of China (2018AAA0100704), NSFC \#61932020, \#62172279,  Science and Technology Commission of Shanghai Municipality (Grant No. 20ZR1436000), and “Shuguang Program" supported by Shanghai Education Development Foundation and Shanghai Municipal Education Commission. 


\newpage

{\small
\bibliographystyle{ieee_fullname}
\bibliography{ref}

\begin{thebibliography}{10}\itemsep=-1pt

\bibitem{Arteta16}
Carlos Arteta, Victor Lempitsky, and Andrew Zisserman.
\newblock Counting in the wild.
\newblock In {\em European Conference on Computer Vision}, 2016.

\bibitem{I3D}
Joao Carreira and Andrew Zisserman.
\newblock Quo vadis, action recognition? a new model and the kinetics dataset.
\newblock In {\em proceedings of the IEEE Conference on Computer Vision and
  Pattern Recognition}, pages 6299--6308, 2017.

\bibitem{chen2017action}
Chen Chen, Baochang Zhang, Zhenjie Hou, Junjun Jiang, Mengyuan Liu, and Yun
  Yang.
\newblock Action recognition from depth sequences using weighted fusion of 2d
  and 3d auto-correlation of gradients features.
\newblock {\em Multimedia Tools and Applications}, 76(3):4651--4669, 2017.

\bibitem{ViT}
Alexey Dosovitskiy, Lucas Beyer, Alexander Kolesnikov, Dirk Weissenborn,
  Xiaohua Zhai, Thomas Unterthiner, Mostafa Dehghani, Matthias Minderer, Georg
  Heigold, Sylvain Gelly, et~al.
\newblock An image is worth 16x16 words: Transformers for image recognition at
  scale.
\newblock {\em arXiv preprint arXiv:2010.11929}, 2020.

\bibitem{RepNet}
Debidatta Dwibedi, Yusuf Aytar, Jonathan Tompson, Pierre Sermanet, and Andrew
  Zisserman.
\newblock Counting out time: Class agnostic video repetition counting in the
  wild.
\newblock In {\em IEEE/CVF Conference on Computer Vision and Pattern
  Recognition (CVPR)}, June 2020.

\bibitem{TANet}
Zhaoyang~Liu et al.
\newblock Tam: Temporal adaptive module for video recognition.
\newblock In {\em ICCV}, 2021.

\bibitem{X3D}
Christoph Feichtenhofer.
\newblock X3d: Expanding architectures for efficient video recognition.
\newblock In {\em CVPR}, 2020.

\bibitem{guo2011simple}
Hongwei Guo.
\newblock A simple algorithm for fitting a gaussian function [dsp tips and
  tricks].
\newblock {\em IEEE Signal Processing Magazine}, 28(5):134--137, 2011.

\bibitem{ResNet}
Kaiming He, Xiangyu Zhang, Shaoqing Ren, and Jian Sun.
\newblock Deep residual learning for image recognition.
\newblock In {\em Proceedings of the IEEE Conference on Computer Vision and
  Pattern Recognition (CVPR)}, June 2016.

\bibitem{huang2020improving}
Yifei Huang, Yusuke Sugano, and Yoichi Sato.
\newblock Improving action segmentation via graph-based temporal reasoning.
\newblock In {\em Proceedings of the IEEE/CVF conference on computer vision and
  pattern recognition}, pages 14024--14034, 2020.

\bibitem{kay2017kinetics}
Will Kay, Joao Carreira, Karen Simonyan, Brian Zhang, Chloe Hillier, Sudheendra
  Vijayanarasimhan, Fabio Viola, Tim Green, Trevor Back, Paul Natsev, et~al.
\newblock The kinetics human action video dataset.
\newblock {\em arXiv preprint arXiv:1705.06950}, 2017.

\bibitem{kitaev2020reformer}
Nikita Kitaev, Łukasz Kaiser, and Anselm Levskaya.
\newblock Reformer: The efficient transformer, 2020.

\bibitem{kobayashi2006three}
Takumi Kobayashi and Nobuyuki Otsu.
\newblock A three-way auto-correlation based approach to human identification
  by gait.
\newblock In {\em IEEE Workshop on Visual Surveillance}, volume~1, page~4.
  Citeseer, 2006.

\bibitem{kobayashi2009three}
Takumi Kobayashi and Nobuyuki Otsu.
\newblock Three-way auto-correlation approach to motion recognition.
\newblock {\em Pattern Recognition Letters}, 30(3):212--221, 2009.

\bibitem{kobayashi2012motion}
Takumi Kobayashi and Nobuyuki Otsu.
\newblock Motion recognition using local auto-correlation of space--time
  gradients.
\newblock {\em Pattern Recognition Letters}, 33(9):1188--1195, 2012.

\bibitem{lengvenis2013application}
Paulius Lengvenis, Rimvydas Simutis, Vygandas Vaitkus, and Rytis
  Maskeli{\=u}nas.
\newblock Application of computer vision systems for passenger counting in
  public transport.
\newblock {\em Elektronika ir Elektrotechnika}, 19(3):69--72, 2013.

\bibitem{synthesized}
Ofir Levy and Lior Wolf.
\newblock Live repetition counting.
\newblock In {\em Proceedings of the IEEE International Conference on Computer
  Vision (ICCV)}, December 2015.

\bibitem{Li_2018_CVPR}
Xiu Li, Hongdong Li, Hanbyul Joo, Yebin Liu, and Yaser Sheikh.
\newblock Structure from recurrent motion: From rigidity to recurrency.
\newblock In {\em Proceedings of the IEEE Conference on Computer Vision and
  Pattern Recognition (CVPR)}, June 2018.

\bibitem{lian2021locating}
Dongze Lian, Xianing Chen, Jing Li, Weixin Luo, and Shenghua Gao.
\newblock Locating and counting heads in crowds with a depth prior.
\newblock {\em IEEE Transactions on Pattern Analysis and Machine Intelligence},
  2021.

\bibitem{lian2019density}
Dongze Lian, Jing Li, Jia Zheng, Weixin Luo, and Shenghua Gao.
\newblock Density map regression guided detection network for rgb-d crowd
  counting and localization.
\newblock In {\em Proceedings of the IEEE/CVF Conference on Computer Vision and
  Pattern Recognition}, pages 1821--1830, 2019.

\bibitem{liu2019context}
Weizhe Liu, Mathieu Salzmann, and Pascal Fua.
\newblock Context-aware crowd counting.
\newblock In {\em Proceedings of the IEEE/CVF Conference on Computer Vision and
  Pattern Recognition}, pages 5099--5108, 2019.

\bibitem{swin-transformer}
Ze Liu, Yutong Lin, Yue Cao, Han Hu, Yixuan Wei, Zheng Zhang, Stephen Lin, and
  Baining Guo.
\newblock Swin transformer: Hierarchical vision transformer using shifted
  windows.
\newblock In {\em Proceedings of the IEEE/CVF International Conference on
  Computer Vision (ICCV)}, pages 10012--10022, October 2021.

\bibitem{video-swin-transformer}
Ze Liu, Jia Ning, Yue Cao, Yixuan Wei, Zheng Zhang, Stephen Lin, and Han Hu.
\newblock Video swin transformer.
\newblock {\em arXiv preprint arXiv:2106.13230}, 2021.

\bibitem{lu2018class}
Erika Lu, Weidi Xie, and Andrew Zisserman.
\newblock Class-agnostic counting.
\newblock In {\em Asian conference on computer vision}, pages 669--684.
  Springer, 2018.

\bibitem{Auto-correlation1}
Costas Panagiotakis, Giorgos Karvounas, and Antonis Argyros.
\newblock Unsupervised detection of periodic segments in videos.
\newblock In {\em 2018 25th IEEE International Conference on Image Processing
  (ICIP)}, pages 923--927. IEEE, 2018.

\bibitem{P3D}
Zhaofan Qiu, Ting Yao, and Tao Mei.
\newblock Learning spatio-temporal representation with pseudo-3d residual
  networks.
\newblock In {\em proceedings of the IEEE International Conference on Computer
  Vision}, pages 5533--5541, 2017.

\bibitem{Pedestrian}
Yang Ran, Isaac Weiss, Qinfen Zheng, and Larry~S. Davis.
\newblock Pedestrian detection via periodic motion analysis.
\newblock {\em Int. J. Comput. Vis.}, 71(2):143--160, 2007.

\bibitem{Ranjan_2018_ECCV}
Viresh Ranjan, Hieu Le, and Minh Hoai.
\newblock Iterative crowd counting.
\newblock In {\em Proceedings of the European Conference on Computer Vision
  (ECCV)}, September 2018.

\bibitem{Ribnick_3dreconstruction}
Evan Ribnick, Nikos Papanikolopoulos, Evan Ribnick, and Nikolaos
  Papanikolopoulos.
\newblock 3d reconstruction of periodic motion from a single view.

\bibitem{Runia_2018_CVPR}
Tom F.~H. Runia, Cees G.~M. Snoek, and Arnold W.~M. Smeulders.
\newblock Real-world repetition estimation by div, grad and curl.
\newblock In {\em Proceedings of the IEEE Conference on Computer Vision and
  Pattern Recognition (CVPR)}, June 2018.

\bibitem{seenouvong2016computer}
Nilakorn Seenouvong, Ukrit Watchareeruetai, Chaiwat Nuthong, Khamphong
  Khongsomboon, and Noboru Ohnishi.
\newblock A computer vision based vehicle detection and counting system.
\newblock In {\em 2016 8th International Conference on Knowledge and Smart
  Technology (KST)}, pages 224--227. IEEE, 2016.

\bibitem{soro2019recognition}
Andrea Soro, Gino Brunner, Simon Tanner, and Roger Wattenhofer.
\newblock Recognition and repetition counting for complex physical exercises
  with deep learning.
\newblock {\em Sensors}, 19(3):714, 2019.

\bibitem{tan2019crowd}
Xin Tan, Chun Tao, Tongwei Ren, Jinhui Tang, and Gangshan Wu.
\newblock Crowd counting via multi-layer regression.
\newblock In {\em Proceedings of the 27th ACM International Conference on
  Multimedia}, pages 1907--1915, 2019.

\bibitem{C3D}
Du Tran, Lubomir Bourdev, Rob Fergus, Lorenzo Torresani, and Manohar Paluri.
\newblock Learning spatiotemporal features with 3d convolutional networks.
\newblock In {\em Proceedings of the IEEE international conference on computer
  vision}, pages 4489--4497, 2015.

\bibitem{vaswani2017attention}
Ashish Vaswani, Noam Shazeer, Niki Parmar, Jakob Uszkoreit, Llion Jones,
  Aidan~N Gomez, {\L}ukasz Kaiser, and Illia Polosukhin.
\newblock Attention is all you need.
\newblock In {\em Advances in neural information processing systems}, pages
  5998--6008, 2017.

\bibitem{Auto-correlation2}
Michail Vlachos, Philip Yu, and Vittorio Castelli.
\newblock On periodicity detection and structural periodic similarity.
\newblock In {\em Proceedings of the 2005 SIAM international conference on data
  mining}, pages 449--460. SIAM, 2005.

\bibitem{Wan_2019_ICCV}
Jia Wan and Antoni Chan.
\newblock Adaptive density map generation for crowd counting.
\newblock In {\em Proceedings of the IEEE/CVF International Conference on
  Computer Vision (ICCV)}, October 2019.

\bibitem{TSN2016ECCV}
Limin Wang, Yuanjun Xiong, Zhe Wang, Yu Qiao, Dahua Lin, Xiaoou Tang, and Luc
  {Val Gool}.
\newblock Temporal segment networks: Towards good practices for deep action
  recognition.
\newblock In {\em ECCV}, 2016.

\bibitem{Zhang_2020_CVPR}
Huaidong Zhang, Xuemiao Xu, Guoqiang Han, and Shengfeng He.
\newblock Context-aware and scale-insensitive temporal repetition counting.
\newblock In {\em IEEE/CVF Conference on Computer Vision and Pattern
  Recognition (CVPR)}, June 2020.

\bibitem{DBLP:journals/corr/abs-2103-13096}
Yunhua Zhang, Ling Shao, and Cees G.~M. Snoek.
\newblock Repetitive activity counting by sight and sound.
\newblock {\em CoRR}, abs/2103.13096, 2021.

\bibitem{MCNN}
Yingying Zhang, Desen Zhou, Siqin Chen, Shenghua Gao, and Yi Ma.
\newblock Single-image crowd counting via multi-column convolutional neural
  network.
\newblock In {\em 2016 IEEE Conference on Computer Vision and Pattern
  Recognition (CVPR)}, pages 589--597, 2016.

\end{thebibliography}
}

\appendix
\setcounter{page}{1}

\twocolumn[
\centering
\Large
\textbf{Encoding Multi-scale Temporal Correlation with Transformers for Repetitive Action Counting} \\
\vspace{0.5em}Supplementary Material \\
\vspace{1.0em}
] 
\appendix

\section{Extra experiments}

\subsection{Density map}

\textbf{Gaussianization}.
Assuming we have the label as $Y=[y1,y2,\dots,y_n]$. A pair of $y_i$ and $y_j$ represents the frames at the beginning and end of a repetition action where $j = i+1$. To calculate the Gaussian function \cite{guo2011simple} with 99\% confidence interval shown below, we need to figure out the mean $\mu$ and the variance $\sigma$. Because of 99\% confidence interval meaning $\mu \pm 3\sigma$, we could get $\mu$ and $\sigma$ by $y_{i,j} = \mu \pm 3\sigma$, where $i$ and $j$ is the pair of the frames. 

 Therefore, through the Gaussian function $g_\sigma(x)$, we can get the probability density distribution $G_\sigma(y)$ from $y_i$ to $y_j$. Then $d_k$ can take the integral by \cref{eq:F_DM}. At last, we could get the predict results of density map $D = [d_1,d_2,\dots,d_n]$.  
 \begin{equation}
     d_k = \int_{y_k - 0.5}^{y_k + 0.5} G_\sigma(y)dy,\quad k\in[i,j]
     \label{eq:F_DM}
\end{equation}

To compare different generate methods of density map, We adjust the mean $\mu$ of the Gaussian function to the beginning frame of one period or the ending and retrain the model which has a merging density map predictor. Then we obtain the output of different positions by adjusting the weight of predict density maps.
\begin{table}[ht]

	\centering
	\begin{tabular}{c|cc}
		\hline
		                     & \multicolumn{2}{c}{RepCount A}                        \\ \hline
		Generate density map & \multicolumn{1}{c|}{MAE$\downarrow$} & OBO $\uparrow$ \\ \hline
		Begin                & \multicolumn{1}{c|}{0.5295}          & 0.2052         \\
		Mid                  & \multicolumn{1}{c|}{0.4936}          & 0.2052         \\
		End                  & \multicolumn{1}{c|}{0.5192}          & 0.192          \\
		Merge                & \multicolumn{1}{c|}{0.5142}          & 0.2009         \\ \hline
	\end{tabular}

	\caption{
		\textbf{The density maps with different mean $\mu$ of the Gaussian function $G$.} \emph{Begin} is means the density map generated with $G$, where the $\mu$ is the beginning frame. Similarly, \emph{End} represents the $\mu$ is the ending frame. \emph{Mid} is as same as our model \textbf{TranRAC}.
	}
	\label{tab:multi density maps}
\end{table}

The result as the \cref{tab:multi density maps} shown, Merging density maps does not give the models better performance. Because the amount of video frames is 64, moving $\mu$ to begin or end will lose the information of the first period or the last. The density map generated by the mean in the mid-frame has the best effort.

\subsection{Sample rate}

We conducted the experiment to verify the impact of adding the number of video frames. Due to our model based on the density map, we use a one-dimensional spatial distribution to represent the distribution of periods in time. 

\begin{table}[ht]
	\begin{center}
		\begin{tabular}{c|cc}

			\hline
			                 & \multicolumn{2}{c}{RepCount A}                         \\ \hline
			Method           & \multicolumn{1}{c|}{MAE$\downarrow$} & OBO  $\uparrow$ \\\hline
			Ours(single)-64  & \multicolumn{1}{c|}{0.6595}          & 0.185           \\
			Ours(single)-128 & \multicolumn{1}{c|}{0.6191}          & 0.191           \\ \hline
		\end{tabular}
	\end{center}
	\caption{
		Experiment results of model with different sample rate when trained on train set of \emph{RepCount part-A}. 64 and 128 indicates different frame sample number from initial videos.
	}
	\label{tab:64_128}
\end{table}

Experimental results show that increasing video frames can improve the performance of density maps to a certain extent (see \cref{tab:64_128}). A better result in terms of MAE error is achieved when we select 128 frames of a video.

\subsection{Scale ablation}

We verify the effect of different scales by building tree pipelines, where the input of Encoder $\phi$ with distinct length video subsequences. As shown in the one to three rows of \cref{tab:scales}, because the different temporal scales of video subsequence extract information in their scale, they have different performances of repetition counting. Concatenating the multi-scale video sequences contributes to capturing different period length actions and brings the model greater robustness.
\begin{table*}[htbp]
\centering
\begin{tabular}{c|cc|cc}
\hline
\multirow{2}{*}{Division}  & \multicolumn{2}{c|}{Regular setting}   & \multicolumn{2}{c}{Open-set setting}               \\ \cline{2-5} 
                        & \multicolumn{1}{c|}{Num. of videos} & Total frames & \multicolumn{1}{c|}{Num. of videos} & Total frames \\ \cline{1-5} 
Train                   & \multicolumn{1}{c|}{758}            & 637,545             & \multicolumn{1}{c|}{655}            &   560,402          \\
val                     & \multicolumn{1}{c|}{131}            & 109,854             & \multicolumn{1}{c|}{130}            &   77,103      \\
test                    & \multicolumn{1}{c|}{152}            & 129,993             & \multicolumn{1}{c|}{256}            &   239,887      \\ \hline
\end{tabular}
\caption{
Regular setting and Open-setting of \emph{RepCount partA}
}
\label{tab:openset}
\end{table*}

\begin{table*}[ht]
\centering
\begin{tabular}{c|cccc}
\hline
\multirow{2}{*}{Method}  & \multicolumn{2}{c|}{regular setting} & \multicolumn{2}{c}{Open-set setting} \\ \cline{2-5} 
                        & MAE$\downarrow$   & \multicolumn{1}{c|}{OBO$\uparrow$ }   & MAE$\downarrow$             & OBO$\uparrow$               \\ \hline
Huang et al. \cite{huang2020improving}   &  0.5267   & \multicolumn{1}{c|}{0.1589}    &    1.0000     &       0.0000           \\
Ours          & 0.4431  & \multicolumn{1}{c|}{0.2913}      &        0.6249 & 0.2040                  \\ \hline
\end{tabular}
\caption{
Performance of different methods on two settings of \emph{RepCount partA}.
}
\label{tab:reviewer3}
\end{table*}

\begin{table}[ht]
\begin{center}

\begin{tabular}{c|cc}

\hline
        & \multicolumn{2}{c}{RepCount A}   \\ \hline
Method  & \multicolumn{1}{c|}{MAE$\downarrow$}    & OBO $\uparrow$    \\ \hline
Scale-1 & \multicolumn{1}{c|}{0.6595} & 0.1854 \\
Scale-4 & \multicolumn{1}{c|}{0.5434} & 0.2649 \\
Scale-8 & \multicolumn{1}{c|}{0.6657} & 0.192  \\
Ours    & \multicolumn{1}{c|}{0.4431} & 0.2913 \\ \hline
\end{tabular}
\caption{
\textbf{The experimental results of pipelines with different scales.} \textit{Scale-i}, where $i\in\{1,4,8\}$, represents the temporal length of video subsequence, which is the input of Encoder $\phi$. The \textit{Ours}, means cancatenating three scales video subsequence together, have the lowest MAE. We build all the above models by extracting 64 frames from the original video.}
\label{tab:scales}
\end{center}
\end{table}

\subsection{Receptive field}
Usually, a more large video subsequence of scale has a more substantial receptive field. And considering the more large sample rate will extract more video information, we compared the different video subsequences of scale at different sample rates. 

As \cref{tab:8+128} indicated, when the sample rate is the little one, 64 frames extracted from one original video, single-frame is as similar as 8-frames. But increasing the sample rate to 128, The performance of 8-frames is far better than that of single-frames.

\begin{table}[ht]
	\begin{center}
		\begin{tabular}{c|ccc}
			\hline
			                     & \multicolumn{3}{c}{RepCount A}                                                         \\ \hline
			Samle rates          & \multicolumn{1}{c|}{Scales}    & \multicolumn{1}{c|}{MAE$\downarrow$} & OBO $\uparrow$ \\ \hline
			\multirow{2}{*}{64}  & \multicolumn{1}{c|}{Scale-1}   & \multicolumn{1}{c|}{0.6595}          & 0.185          \\
			                     & \multicolumn{1}{c|}{Scale-8}   & \multicolumn{1}{c|}{0.6657}          & 0.192          \\ \hline
			\multirow{2}{*}{128} & \multicolumn{1}{c|}{Scale-1}   & \multicolumn{1}{c|}{0.6191}          & 0.191          \\
			                     & \multicolumn{1}{c|}{Scale-8}   & \multicolumn{1}{c|}{0.4926}          & 0.2302         \\ \hline
		\end{tabular}
		\caption{
			\textbf{Performance of different scales at different sample rates.} The first column, \emph{sample rates}, indicates different frame sample number from initial videos. \emph{Scale-i}, where $i\in\{1,8\}$, represents the temporal length of video subsquence.
		}
		\label{tab:8+128}
	\end{center}
\end{table}

We believe that there is an optimal scale of video subsequence for the same dataset under the same sampling rate. Due to the operation of sampling video to the fixed number of frames, the duration of per repetitive action will be shorter with the decrease of the sample rate. Large-scale video sequences will lose their advantages when the sample rate is 64 frames per video because of the shorter duration. But when the sampling rate increases to 128, there is more difference reflected between different video subsequence scales.  Although single-frame has improved the MAE and OBO, the progress of 8-frames is more excellent.

\subsection{Compare to action segmentation}

We elaborate the definitions and differences between action segmentation and repetitive action counting. Given an input video, action segmentation is to segment the temporal bound for different types of actions but repetitive action counting aims to count the number of repetitive action \cite{RepNet,Zhang_2020_CVPR}. Two main differences are as follows: i) for action segmentation, the same action continuously repeating many times will be segmented into a single temporal bound. Thus, it is difficult to handle videos with high-frequency repetitive actions. However, the variation in the frequency of repetitive action is huge, \emph{e.g.}, the min/max cycle length is 0.1/10.96 in our dataset, which degenerates action segmentation method; ii) action segmentation can only address predefined action types and cannot handle open-set setting, where action types in the test set that do not exist in the training set. However, repetitive counting is to record repeated actions regardless of the action category. 

To verify the above differences, we further perform experiments in \cref{tab:reviewer3}. Under both settings of \cref{tab:openset}, our method achieves better performance  compared to action segmentation \cite{huang2020improving}. Therefore, our task is not a trivial case of action segmentation.

\section{Dataset description}
\subsection{Data duration}
The duration in Table 1 means video length, not the cycle length of each action, which shows that our dataset contains longer videos. In addition, not only the short action but the diversity of actions makes the task harder and more useful. The min/max cycle length between Ours and UCF526\cite{Zhang_2020_CVPR} is (0.1/10.96 vs 0.12/6.76), which shows our dataset is more challenging. 

\subsection{Open-set setting}

We add an open-set setting to demonstrate the better ability of our method when dealing with unseen action types in the training set. Therefore, we re-split the \emph{RepCount partA} into a new train/val and test subset. For regular settings, videos are divided randomly. For Open-set setting, the action types in train/val/test are disjoint, where the actions in the test set do not appear in the training set. More details show in \cref{tab:openset}.

\end{document}